\documentclass[11pt, a4paper, copyright, goog]{google}

\usepackage[numbers, sort&compress]{natbib}

\usepackage{ifthen}
\usepackage{makecell}
\usepackage{nicefrac}
\usepackage[most]{tcolorbox}
\usepackage{ragged2e}
\usepackage{setspace}
\usepackage{subcaption}
\usepackage{float}
\usepackage{listings}

\newboolean{ANONYMIZE}
\setboolean{ANONYMIZE}{false}

\usepackage{titlesec}
\titlespacing*{\section}{0pc}{2.76ex plus 3.7pt minus 2.8pt}{4.6pt}
\titlespacing*{\subsection}{0pc}{2.3ex plus 2.76pt minus 1.84pt}{1.84pt}
\titlespacing*{\subsubsection}{0pc}{1.84ex plus 2.3pt minus 1.4pt}{1.84pt}

\paperurl{}
\renewcommand{\today}{}

\uselogo{}

\title{JAXBench: Benchmarking Autonomous TPU Kernel Optimization}

\correspondingauthor{aryatschand@g.harvard.edu, sethusankaran@google.com}

\author[*,1,4]{Arya Tschand}
\author[*,2,4]{Charles Hong}
\author[3]{Julian Walker}
\author[4]{Nina Cai}
\author[4]{Shangkun Wang}
\author[4]{Suvinay Subramanian}
\author[4]{Sundar Dev}
\author[1]{Vijay Janapa Reddi}
\author[3]{Amir Yazdanbakhsh}
\author[4]{Sethu Sankaran}

\affil[*]{Equal contributions, work done while at Google}
\affil[1]{Harvard University}
\affil[2]{UC Berkeley}
\affil[3]{Google DeepMind}
\affil[4]{Google}

\begin{abstract}
\vspace{-1.2em}
{\normalfont\mdseries\noindent Code: \url{https://github.com/AI-Hypercomputer/accelerator-agents/tree/main/JAXBench}}\\[1.2em]
Rigorous benchmarks have driven progress in autonomous GPU kernel performance optimization by establishing a shared target to hillclimb on, but no equivalent exists for TPUs. We present \textsc{JAXBench}, a TPU-native benchmark suite for AI-generated kernel optimization on Google Cloud TPUs. \textsc{JAXBench} comprises 50 JAX workloads that are both \textit{relevant} and provide \textit{headroom} for optimization. We extract 17 production ML operators from architectures in the public MaxText library such as Llama-3.1, DeepSeek-V3, Mixtral, Mamba-2, and AlphaFold2, and translate 33 operators from KernelBench that are validated for correctness and set with new problem sizes that achieve high TPU v6e MXU utilization. Eight of the 17 production operators ship with hand-optimized Pallas kernels from the public Tokamax library and block-size tuned to establish an expert upper-bound baseline. We evaluate four feedback-driven methods on generating candidate Pallas kernels for \textsc{JAXBench}. Across the full suite with Gemini 3 Flash, we find that target-specific context matters more than model scale on a sparsely-documented DSL like Pallas. Conditioning on curated TPU documentation raises per-sample correctness from 5.8\% to 37.3\% and solves 48 of 50 benchmarks at a $1.28\times$ geomean speedup. Search structure yields significant gains once correctness is achieved, with Autocomp's beam-search pipeline reaching a $1.36\times$ geomean speedup over XLA. On the 8 hand-tuned kernels, Autocomp reaches $1.60\times$ geomean over XLA, recovering most of the $2.08\times$ Tokamax upper bound but trailing on the specialized paged and ragged attention operators. High-quality TPU kernel optimization remains a challenging task, and we release the \textsc{JAXBench} benchmark, evaluation harness, and baseline results to support open source contributions.
\end{abstract}

\begin{document}

\maketitle

\section{Introduction}

Efficient kernel implementations are a bottleneck in achieving the full potential of hardware accelerators for machine learning. Libraries such as cuBLAS, CUTLASS, and Triton~\cite{cutlass, tillet2019triton} and specialized kernels for GPU~\cite{dao2022flashattention} and TPU~\cite{jiang2026ragged} have unlocked generation-over-generation performance gains, but each new model architecture, quantization scheme, and hardware revision demands fresh low-level implementations. To realize the performance gains enabled by architecture, kernels must be rapidly co-designed between the hardware-specific features and emerging workload characteristics.

\definecolor{jbblue}{RGB}{37,99,180}
\definecolor{jbgreen}{RGB}{46,125,50}
\definecolor{jbpurple}{RGB}{107,70,193}

\begin{figure*}[t]
    \centering
    \includegraphics[width=\textwidth]{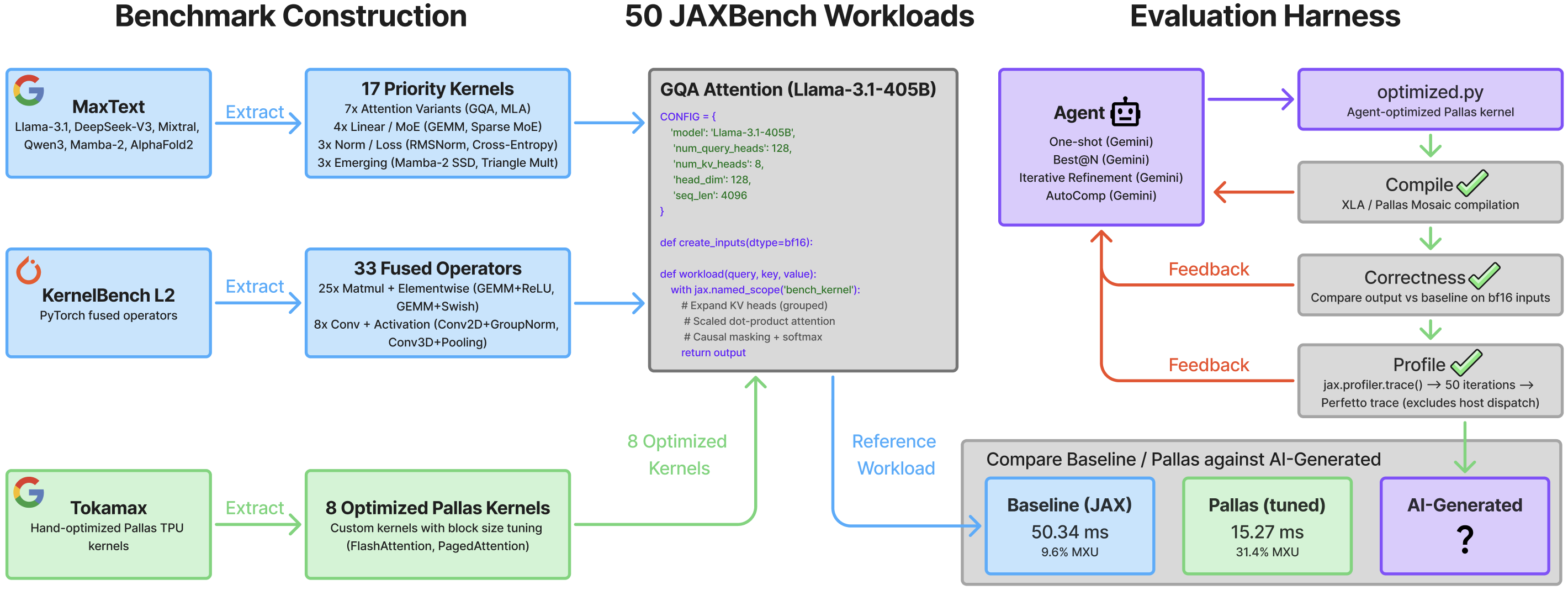}
    \caption{\textbf{JAXBench overview: a TPU-native benchmark for evaluating AI-generated Pallas kernels against hand-tuned baselines on TPU v6e.}
    We construct \textcolor{jbblue}{50 JAX reference workloads (blue)} with 17 priority kernels from MaxText~\cite{maxtext2022} LLMs and 33 fused operators adapted from KernelBench L2, all sized to saturate TPU v6e MXUs. We also extract \textcolor{jbgreen}{8 priority kernels of hand-tuned Pallas implementations (green)} from Tokamax as expert upper bounds. The \textcolor{jbpurple}{agent evaluation harness (purple)} enables evaluation of any LLM-driven method to generate candidate Pallas kernels, which are compiled, correctness-checked on bf16 inputs, and profiled via \texttt{jax.profiler}, with feedback fed back and final kernels compared against both baselines.}
    \label{fig:methodology}
\end{figure*}

This gap has motivated a fast-growing body of work on using language models to generate and optimize kernels automatically, ranging from one-shot completions to iterative agents with compile and profile feedback to evolutionary search over program space~\cite{liao2025kernelevolve, tschand2025swizzleperf, lange2025towards, cao2026k}.
Rigorous benchmarks have been central to this progress.
KernelBench~\cite{ouyang2025kernelbench} standardized the evaluation protocol with 250 PyTorch workloads and has since been extended to Triton, CuTe, and other DSLs.
TritonBench~\cite{li2025tritonbench} specifically targets Triton generation with hardware-aware performance measurement on NVIDIA and AMD GPUs.
FlashInfer-Bench~\cite{xing2026flashinfer} grounds evaluation in real LLM serving traces with expert-written FlashInfer kernels as references.
Together, these benchmarks have driven rapid community progress on correctness and speedup against GPU baselines and have enabled meaningful comparison of methods from best-of-$N$ sampling to reinforcement-learned coding agents.

However, \emph{no analogous benchmark exists for TPU kernel optimization.} Google's Tensor Processing Units~\cite{jouppi2017datacenter} differ fundamentally from GPUs. TPUs are sequential machines with wide SIMD vector registers (8$\times$128 for 32-bit values on v6e) and dedicated 256$\times$256 systolic Matrix Multiply Units (MXUs), rather than the massively parallel SIMT execution model of GPUs. Programming TPUs also requires a distinct software stack. JAX~\cite{bradbury2021jax} programs compile through XLA, and low-level kernel authoring goes through Pallas, which lowers to the Mosaic backend rather than Triton. Pallas kernels must reason about TPU-specific concerns including VMEM/SMEM/HBM memory hierarchies, software pipelining with prefetch scheduling, block shape constraints, and the lexicographic grid traversal order that Mosaic enforces~\cite{jax_pallas_tpu}. Pallas also appears in LLM training data orders of magnitude less frequently than CUDA or even Triton, so models that fluently write GPU kernels routinely hallucinate Pallas APIs, emit memory-space annotations that do not type-check, or violate systolic tiling constraints in ways that no amount of generic compile feedback can resolve. GPU benchmarks therefore cannot be directly applied because workloads, problem sizes, programming abstractions, and optimization strategies are all GPU-specific.

Recent concurrent work has addressed multi-platform evaluation. MultiKernelBench~\cite{wen2025multikernelbench} extends KernelBench to CUDA, Huawei AscendC, and Google Pallas, finding that the best model achieves only 8.4--10.5\% Pass@1 on Pallas tasks. However, MultiKernelBench does not target JAX, evaluates on TPU v2-8, translates PyTorch workloads at problem sizes too small to saturate the MXU, and does not include production LLM operators or optimized reference kernels. At small shapes, workloads are dominated by memory traffic and launch overhead rather than compute, so any measured speedup reflects bookkeeping rather than genuine algorithmic or scheduling improvement. A useful TPU benchmark must instead push workloads into the compute-bound regime where the MXU is binding and tiling, pipelining, and layout choices actually affect throughput. Table~\ref{tab:related-work} summarizes how existing benchmarks position on these axes. There remains a need for a TPU-native benchmark that (i) uses contemporary TPU hardware, (ii) includes workloads representative of modern LLM training and inference, (iii) provides strong baselines through both XLA compilation and hand-optimized Pallas kernels, and (iv) sizes problems to be compute-bound so optimization headroom is meaningful.

We introduce \textsc{JAXBench}, a benchmark suite of 50 JAX workloads designed specifically for TPU kernel optimization. Our contributions are as follows.

\begin{enumerate}
\item \textbf{TPU-native benchmark suite with 50 relevant workloads.} \textsc{JAXBench} is a benchmark of 17 production operators from real LLM architectures (Llama-3.1, DeepSeek-V3, Mixtral, Mamba-2, AlphaFold2) extracted from MaxText, and 33 fused operator sequences from KernelBench translated from PyTorch to JAX and set with problem sizes adjusted to achieve high TPU MXU utilization. For 8 priority kernels, we additionally provide expert-optimized Pallas TPU kernels from the upstream Tokamax library with tuned block sizes.
\item \textbf{Rigorous and reproducible evaluation harness.} Device-side profiling via \texttt{jax.profiler} Perfetto traces eliminates host dispatch overhead, yielding reproducible per-iteration kernel timings that can be used to easily evaluate any agent framework on \textsc{JAXBench}.
\item \textbf{Agent evaluation and failure modes.} We deploy \textsc{JAXBench} on TPU v6e (Trillium) and evaluate best-of-$N$, iterative feedback agent loops, TPU context-conditioned agent loops, and Autocomp~\cite{hong2025autocomp} augmented with TPU architecture documentation. We also provide insights into failure modes and opportunities for further contributions in TPU-centric kernel optimization.
\end{enumerate}

\begin{table}[t]
\centering
\small
\setlength{\tabcolsep}{5pt}
\caption{Comparison of LLM kernel-generation benchmarks. \textsc{JAXBench} is the only suite targeting contemporary TPUs with production-scale workloads, hardware-saturating problem sizes, and expert-written Pallas reference kernels for a subset.}
\label{tab:related-work}
\begin{tabular}{lllllc}
\toprule
\textbf{Benchmark} & \textbf{Target} & \textbf{\# Tasks} & \textbf{Workload Source} & \textbf{Expert Refs} & \textbf{Sat. Sizes} \\
\midrule
KernelBench~\cite{ouyang2025kernelbench}
  & CUDA / GPU                  & 250 & PyTorch modules    & ---        & Yes     \\
TritonBench~\cite{li2025tritonbench}
  & Triton / GPU                & 184 & Real repos         & Triton     & Partial \\
FlashInfer-Bench~\cite{xing2026flashinfer}
  & CUDA / GPU                  & --  & LLM serving traces & FlashInfer & Yes     \\
MultiKernelBench~\cite{wen2025multikernelbench}
  & Pallas / TPU v2-8 & 285 & PyTorch modules    & ---        & No      \\
\midrule
\textbf{\textsc{JAXBench} (ours)}
  & \textbf{Pallas / TPU v6e}
  & \textbf{50}
  & \makecell[l]{\textbf{Production LLMs +}\\\textbf{Standard JAX}}
  & \textbf{Pallas}
  & \textbf{Yes} \\
\bottomrule
\end{tabular}
\end{table}
\vspace{-1em}

\section{Methodology}

\subsection{Design Principles}
 
\textsc{JAXBench} is built around three principles that together make the benchmark both practically relevant and technically rigorous for TPU kernel optimization.

\paragraph{Relevant workloads.}
The benchmark must evaluate performance on a diverse set of operators representative of real-world TPU customers. We include both well-studied operators where expert-optimized Pallas kernels already exist (e.g., flash attention, grouped-query attention) and emerging operators where no such Pallas kernels have been developed (e.g., Mamba-2 state space duality, AlphaFold2 triangle multiplication). This combination tests automated optimization methods across varying levels of prior human effort.

\begin{figure*}[t]
    \centering
    \begin{subfigure}[t]{0.47\textwidth}
        \centering
        \includegraphics[width=\textwidth]{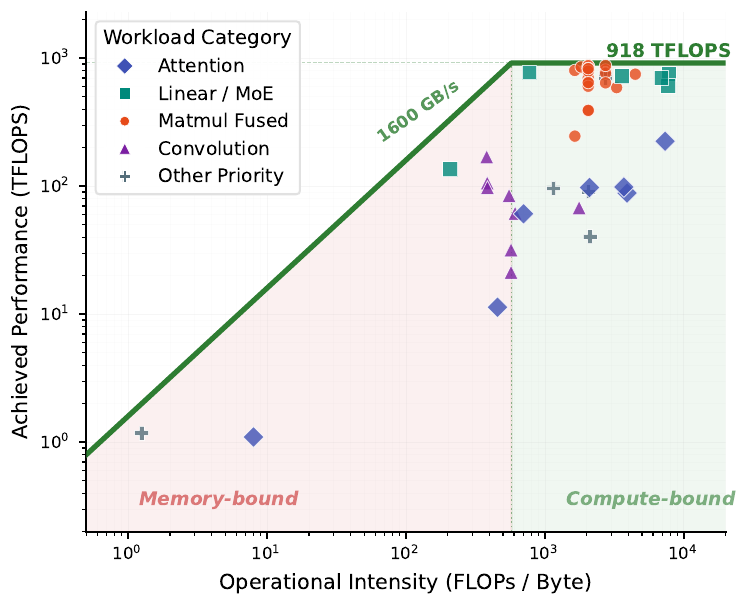}
        \caption{TPU v6e roofline~\cite{williams2009roofline} with all 50 workloads.}
        \label{fig:roofline}
    \end{subfigure}
    \hfill
    \begin{subfigure}[t]{0.52\textwidth}
        \centering
        \includegraphics[width=\textwidth]{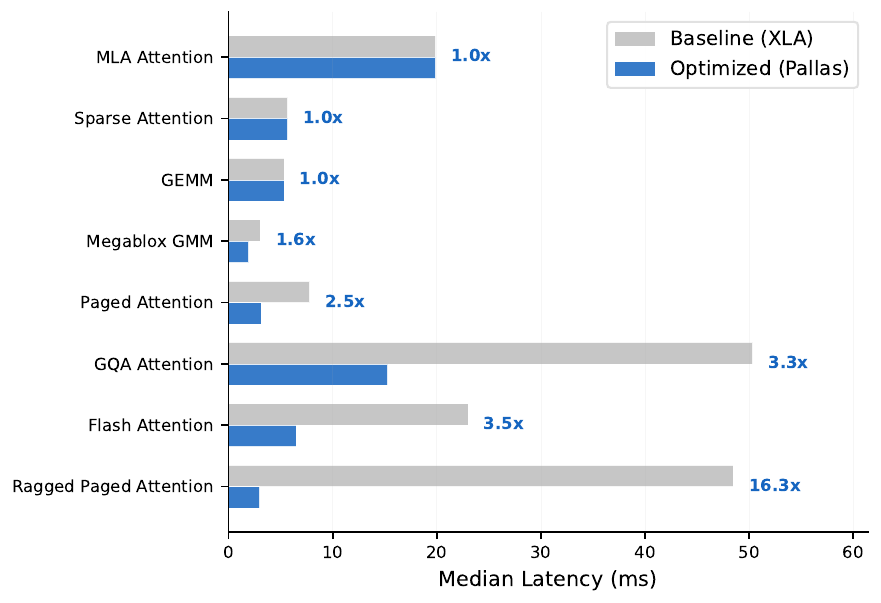}
        \caption{Tokamax Pallas kernel speedup over XLA.}
        \label{fig:speedup}
    \end{subfigure}
    \caption{\textbf{\textsc{JAXBench} workloads are predominantly compute-bound on the TPU v6e roofline, and show headroom where existing Pallas kernel optimization can achieve up to a 16$\times$ speedup over the XLA baseline.}
    Achieved performance vs.\ operational intensity on a single TPU v6e chip (918 TFLOPS bf16, 1640 GB/s HBM) reveals that matmul-fused and linear/MoE kernels already reach 60--95\% MXU utilization under XLA,
    while attention and elementwise kernels remain memory-bound.
    Hand-optimized Pallas kernels close this gap for the 8 priority workloads by tiling computation to reduce redundant HBM traffic, achieving speedups of up to 16.3$\times$ on Ragged Paged Attention and
    3.5$\times$ on Flash Attention. Compute-bound workloads like GEMM show no improvement, confirming that XLA already saturates the MXU for high-OI operations and that optimization effort is best directed at memory-bound kernels.}
    \label{fig:roofline-speedup}
\end{figure*}

\paragraph{Headroom to optimize.}
Workloads and problem sizes are chosen so that the XLA-compiled baseline already achieves high TPU utilization, leaving kernel optimization to compete on genuine algorithmic and scheduling improvements rather than on closing artificial gaps.
This choice reflects how TPUs are actually used. Customers running production workloads tune problem dimensions to maximally utilize the hardware, and FLOPs utilization at the workload level is tightly correlated with FLOPs utilization of the underlying operations. A benchmark that evaluates kernels at undersized, memory-bound, or launch-overhead-bound shapes would not reflect the regime in which TPU users (and therefore TPU kernel optimizers) actually operate. We therefore size each workload to push the relevant operations toward the compute-bound regime, where the MXU is the binding resource and optimization headroom is meaningful. As shown in Figure~\ref{fig:roofline}, matmul-heavy fused operators use dimensions such as $(4096, 8192) \times (8192, 8192)$ in bf16, achieving 60--95\% MXU utilization, and priority kernels use production-scale dimensions from their source architectures. The Llama-3.1-70B GEMM workload, for example, runs at $8192 \times 8192 \times 28672$ and achieves $\sim$79\% MXU utilization. Problem size is itself an axis of optimization, since different shapes expose different opportunities and sufficiently small problems offer no headroom at all. 
 
\paragraph{Reproducible baselines.}
Every workload is implemented in idiomatic JAX, compiled via XLA's \texttt{jax.jit}. This is the strongest reproducible baseline for TPU because XLA already performs aggressive whole-program optimization including operator fusion, buffer assignment, and layout optimization.
For priority kernels, when available, we additionally provide Pallas-optimized variants that represent the current state of the art in hand-tuned TPU kernel performance, establishing an upper bound that automated methods should aspire to match.
 
\subsection{Workload Construction}
\label{sec:workloads}
 
\paragraph{Priority kernels (17 workloads).}
We extract 17 compute-critical operators from production LLM architectures as implemented in MaxText~\cite{maxtext2022}. The set spans attention variants (flash~\cite{dao2022flashattention}, GQA~\cite{grattafiori2024llama}, MLA~\cite{liu2024deepseek}, sparse/splash~\cite{jiang2024mixtral}, flex, paged~\cite{kwon2023efficient}, and ragged paged), dense and sparse linear algebra (GEMM, SwiGLU MLP~\cite{shazeer2020glu}, sparse MoE~\cite{shazeer2017outrageously}, Megablox GMM, ragged dot), normalization and loss (RMSNorm~\cite{zhang2019root}, cross-entropy), and emerging architectures (RetNet retention~\cite{sun2023retentive}, Mamba-2 SSD~\cite{gu2023mamba}, AlphaFold2 triangle multiplication~\cite{jumper2021highly}). Each workload uses production-scale dimensions from its source model. The GQA attention workload, for example, uses Llama-3.1-405B dimensions with 128 query heads and 8 key-value heads at sequence length 4096.
 
Whenever available, we source optimized implementations by importing equivalent Pallas kernels from the upstream JAX Pallas ops library (Tokamax)~\cite{tokamax2024}, then tune their block size parameters via exhaustive grid search on TPU v6e. Eight such high quality kernels were available in the library. The tuning procedure evaluated 203 configurations total, yielding speedups of up to 2.79$\times$ (Megablox GMM) over the Pallas default parameters.

\paragraph{KernelBench fused operators (33 workloads).}
We translate a curated subset of 33 workloads from KernelBench Level~2~\cite{ouyang2025kernelbench} from PyTorch to equivalent JAX. We focus on Level~2 because Level~1 consists of isolated single operators with no fusion structure, whereas Level~2 combines a matrix multiplication or convolution with elementwise operations (activations, normalization, pooling), directly representing the kernel fusion opportunities that are most relevant for TPU optimization. To select a non-redundant subset, we group all Level~2 tasks by their fusion signature, namely the (matmul or convolution) $\times$ (activation, normalization, pooling) operator class they instantiate, and retain one representative per class. This procedure deduplicates structurally identical tasks and yields the 33 workloads we include.

Translation is performed by prompting Gemini with each reference PyTorch module and asking for an idiomatic JAX equivalent. To confirm numerical correctness, we execute the original PyTorch reference and the generated JAX implementation on TPU with matched bf16 inputs and require their outputs to agree under \texttt{jnp.allclose} with $\mathrm{atol} = \mathrm{rtol} = 10^{-2}$, the standard bf16 tolerance. Any workload that fails this check is regenerated or repaired by hand before inclusion.

Many of the original KernelBench problem sizes are too small to saturate TPU MXUs, since the 256$\times$256 systolic arrays require large matrix dimensions to achieve high utilization. Rather than apply a uniform scaling factor, we tune each workload's free dimensions independently by sweeping over candidate shapes and freezing the smallest configuration whose XLA baseline reaches at least 60\% MXU utilization on TPU v6e. This per-operator procedure yields a heterogeneous set of shapes (e.g., batch size 4096 with feature dimensions $8192 \times 8192$ in bf16 for matmul-fused operators), all of which are compute-bound rather than launch-overhead-bound on TPU.
 
\paragraph{Workload interface.}
Every workload module exposes a standardized interface composed of a \texttt{CONFIG} dictionary of hyperparameters, a \texttt{create\_inputs(dtype)} function that generates input tensors in bf16, and a \texttt{workload(*inputs)} function containing the computation to benchmark, which is annotated with \texttt{jax.named\_scope} by the profiler harness for visibility.
The benchmark runner JIT-compiles each workload via \texttt{jax.jit} and executes it under controlled timing (Section~\ref{sec:timing}).
 
\subsection{Timing and Profiling}
\label{sec:timing}

Accurate kernel-level timing on TPUs requires device-side measurement.
Wall-clock timing via \texttt{time.perf\_counter()} with \texttt{block\_until\_ready()} includes host-side overhead from Python dispatch, JAX runtime scheduling, and synchronization.
For short kernels ($<\!1$\,ms), this overhead can constitute 10--20\% of measured time.

We use \texttt{jax.profiler.trace()} to capture Perfetto-compatible traces and extract per-iteration execution times. For each workload, we (1) create inputs in bf16, (2) JIT-compile, (3) run 5 warmup iterations, (4) execute 50 timed iterations under the profiler, and (5) parse the trace to extract \texttt{jit\_*()}-level device events.
These events capture the total device-side execution time per iteration, regardless of how XLA or Pallas decomposes the computation internally (single fusion, multiple fusions, or a Pallas kernel call).
We report the median over 50 iterations as the primary metric.
  
\paragraph{FLOP counting and utilization.}
We measure MXU utilization on TPU v6e and operational intensity of ragged paged attention~\cite{jiang2026ragged} as
\begin{equation}
\text{MXU util}_{\text{v6e}} = \frac{\text{FLOPs} / t_{\text{median}}}{918 \times 10^{12}},
\qquad
\text{OI}_{\text{ragged-attn}} = \frac{\text{FLOPs}_{\text{v6e}}}{\text{MemBW}_{\text{v6e}}} = \frac{G \sum_{i=1}^{B} L_i}{GB + \sum_{i=1}^{B} L_i}
\end{equation}
where 918 TFLOPS is the TPU v6e single-chip bf16 peak, $G = H_q / H_{kv}$ is the GQA grouping factor, $B$ is the number of sequences, and $\{L_i\}$ are the per-sequence KV lengths. The numerator counts attention FLOPs, which scale as $H_q \sum_i L_i$ since every query head attends over its full KV sequence. The denominator counts bytes read, with $H_q B$ from the per-sequence query tensors and $H_{kv} \sum_i L_i$ from the shared K and V cache (with constants for $d_{\text{head}}$ and bytes-per-element absorbed into the proportionality). Dividing numerator and denominator by $H_{kv}$ yields the form above. In the long-context limit $\sum_i L_i \gg GB$ the operational intensity therefore approaches $G$, reflecting that each KV token is reused by exactly $G$ query heads and that this reuse caps achievable OI regardless of sequence length. For Llama-3.1-405B ($G = 16$), this upper bound of 16 FLOP/byte sits far below the v6e ridge point of $\mathrm{OI} \approx 560$, confirming that ragged paged attention is memory-bandwidth-bound at any context length.
For the workloads, FLOPs are extracted from XLA's \texttt{cost\_analysis()} on the compiled HLO program and bytes accessed are computed analogously for each operator class.

\section{Results}
\label{sec:results}

\subsection{Evaluation Protocol}
\label{sec:eval}

We evaluate generated kernels along three axes. The first two are gating criteria, and the third is our reported performance metric. We additionally report one derived budget metric (fast$_1$@$N$) to summarize how quickly each method produces useful kernels.

\paragraph{1.\ Compilability.} Does the generated code compile to a valid JAX/Pallas program? For Pallas kernels, this requires correct use of \texttt{BlockSpec}, \texttt{PrefetchScalarGridSpec}, memory space annotations, and block shape constraints specific to TPU.

\paragraph{2.\ Correctness.} Does the generated kernel produce outputs that match the reference implementation? We check numerical agreement on randomized bf16 inputs under \texttt{jnp.allclose} with $\mathrm{atol} = \mathrm{rtol} = 10^{-2}$, the standard bf16 tolerance. Kernels that fail either gate contribute $1\times$ speedup to the aggregate.

\paragraph{3.\ Speedup over XLA.} Our primary performance metric is wall-clock speedup of the generated kernel over the XLA-compiled JAX baseline, measured with device-side profiler timing (Section~\ref{sec:timing}). For the 8 priority kernels with hand-optimized Pallas implementations, we report performance against those references separately as an upper-bound comparison rather than as the denominator of the aggregate speedup.

\paragraph{fast$_1$@$N$.} As a sample-efficiency metric, fast$_1$@$N$ reports the fraction of benchmarks whose best-so-far generated kernel already beats the XLA baseline after $N$ cumulative samples emitted by the method. It complements geomean speedup by measuring breadth of coverage rather than magnitude of improvement at a given budget.

\paragraph{Evaluated methods.} We evaluate four approaches.

(1)~\textbf{Best-of-$N$ generation.} Each sample is an independent one-shot completion conditioned on a TPU v6e preamble and the JAX source, and the best correct sample is selected. See Appendix~\ref{app:prompts}.

(2)~\textbf{Iterative refinement.} Agent loops with access to compilation errors, correctness results, and profiler summaries as feedback between turns, following the paradigm of KernelBench~\cite{ouyang2025kernelbench} but adapted for JAX/Pallas. We run 18 independent chains of 8 turns each.

(3)~\textbf{Iterative refinement with Autocomp context.} Identical to iterative refinement except that every turn receives Autocomp's agent context (hardware architecture summary, per-benchmark-selected Pallas API reference, selected code examples, and rules block) prepended to the preamble. This ablation isolates the effect of curated documentation from the effect of structured search.

(4)~\textbf{Autocomp}~\cite{hong2025autocomp}. An open-source LLM-driven kernel optimizer that ingests publicly available JAX Pallas and Cloud TPU documentation and distills it into four artifacts prepended to every agent prompt: a hardware-architecture summary, a curated Pallas API reference, annotated code examples, and correctness rules. We run a two-phase pipeline: a \emph{translation phase} (4 beam-search iterations) rewrites the XLA baseline into a Pallas kernel, and an \emph{optimization phase} (4 iterations) refines it for performance. Both phases use beam size 3 with 6 new candidates per beam element per iteration. We select Autocomp as our primary agent baseline because it is the most hardware-aware open method available. Other methods such as AlphaEvolve~\cite{novikov2025alphaevolve} and KernelEvolve~\cite{liao2025kernelevolve} are not designed with TPU-specific context, and we leave their evaluation to future work.

\paragraph{Setup.} We evaluate all four methods on TPU v6e (Trillium), starting from the XLA baseline. Best-of-$N$ and both iterative variants use a fixed budget of 144 samples per benchmark. Autocomp uses a beam search of comparable total budget split across translation and optimization phases, and early-stops once its beam stops improving, so its actual sample count per benchmark can be lower than 144. For each method and benchmark we track best-so-far speedup over XLA at every sample, and aggregate across benchmarks by geometric mean.\footnote{We set the speedup floor to $1\times$ to prevent correct but slow kernels from counting as worse than incorrect kernels.} All methods use Gemini 3 Flash~\cite{team2023gemini} for the main evaluation on all 50 benchmarks (Section~\ref{sec:results-full}). We also show an ablation on a 5-kernel subset with Gemini 3.1 Pro (Section~\ref{sec:results-ablation}).

\subsection{Baseline Comparison (Gemini 3 Flash)}
\label{sec:results-full}

We evaluate all four methods across the 50-workload JAXBench suite with Gemini 3 Flash (Table~\ref{tab:flash-main}).

\begin{table}[h]
\centering
\small
\caption{Cross-method comparison on the full 50-benchmark JAXBench suite with Gemini 3 Flash. Per-benchmark best speedups are floored at $1\times$ before aggregating; missing (incorrect) benchmarks count as $1\times$. Per-sample correctness is reported separately in Figure~\ref{fig:failures}.}
\label{tab:flash-main}
\vspace{0.5em}
\begin{tabular}{lccc}
\toprule
Method & Geomean speedup & Mean speedup & Per-benchmark correctness \\
\midrule
Best-of-$N$ & 1.01$\times$ & 1.02$\times$ & 13/50 \\
Iterative refinement & 1.18$\times$ & 1.50$\times$ & 32/50 \\
Iterative + context & 1.28$\times$ & 1.55$\times$ & \textbf{48/50} \\
Autocomp & \textbf{1.36}$\times$ & \textbf{1.70}$\times$ & 45/50 \\
\bottomrule
\end{tabular}
\end{table}

\begin{figure}[h]
\centering
\includegraphics[width=\linewidth]{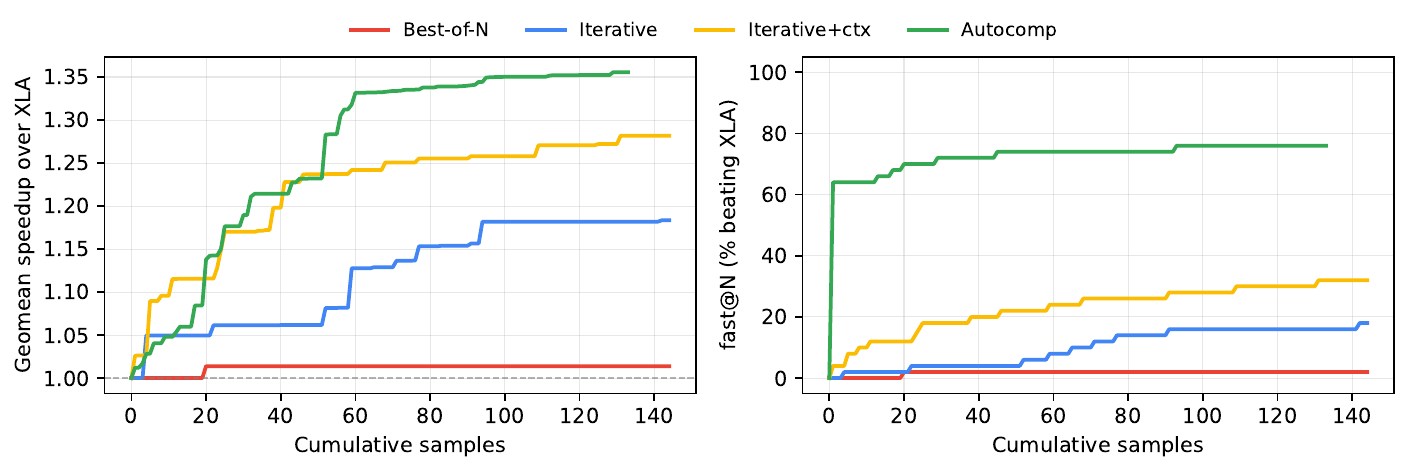}
\caption{Speedup and solved-rate trajectories on the full JAXBench suite (Gemini 3 Flash) as a function of cumulative samples. \textbf{Left:} geomean speedup over XLA (per-kernel best speedups floored at $1\times$ before the geomean, so curves are monotonically non-decreasing). \textbf{Right:} fast$_1$@$N$, the fraction of benchmarks whose best-so-far sample already beats XLA at budget $N$.}
\label{fig:trajectory-flash}
\end{figure}

\paragraph{Speedup trajectory.}
Figure~\ref{fig:trajectory-flash} shows best-so-far geomean speedup (left) and fast$_1$@$N$ (right) as samples accumulate. The method ranking is stable almost from the first sample. Autocomp pulls ahead within the first dozen samples and holds the lead through the full budget, finishing at $1.36\times$ geomean with $76\%$ of benchmarks beating XLA. Iterative+context is the runner-up at $1.28\times$ and $32\%$ fast$_1$@$N$, a clear improvement over plain iterative ($1.18\times$, $18\%$). Best-of-$N$ never climbs meaningfully, producing correct kernels for 13 of 50 benchmarks but only beating XLA on one.

Iterative+context lands \emph{more} correct kernels than Autocomp (48/50 vs.\ 45/50) yet reaches a lower geomean speedup. The two methods spend their 144-sample budget differently: iterative+context debugs until nearly every chain is correct, so it eventually solves 3 of the 5 benchmarks on which Autocomp generates no correct Pallas. Autocomp commits 4 translation cycles up front and then spends its remaining budget on optimization rather than debugging, which enables higher speedups. Per-benchmark best speedups are reported in Figure~\ref{fig:per-benchmark} (Appendix~\ref{app:per-benchmark}).

\begin{figure}[h]
\centering
\includegraphics[width=0.7\linewidth]{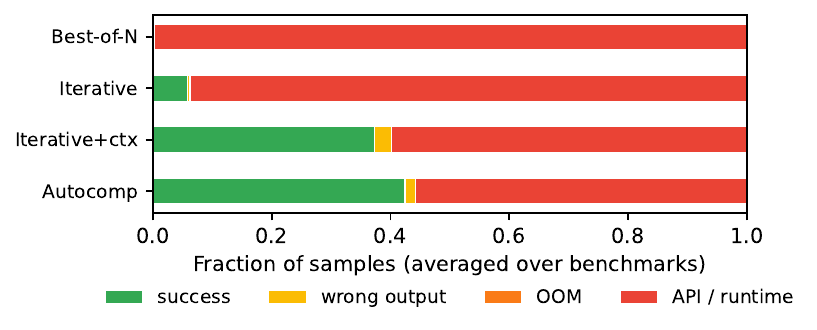}
\caption{Per-sample outcome breakdown on \textsc{JAXBench} (Gemini 3 Flash), pooling all samples across the 50 benchmarks. ``Wrong output'' denotes samples that compile and run but fall outside numerical tolerance. ``API/runtime'' covers unsupported Pallas API syntax and compile/runtime exceptions.}
\label{fig:failures}
\end{figure}

\paragraph{Failure analysis.}
To understand \emph{where} each method spends its sample budget, we classify every sample by outcome (Figure~\ref{fig:failures}). Adding TPU-specific documentation sharply reduces API misuse. Iterative refinement with Autocomp's context raises per-sample correctness from $5.8\%$ to $37.3\%$ with no change to the search algorithm. Pallas API misuse dominates context-free failures, with $99.7\%$ of best-of-$N$ and $93.8\%$ of iterative samples failing at compile or first execution by hallucinating Pallas APIs or misusing \texttt{pallas\_call} arguments and \texttt{BlockSpec} types. API misuse remains a challenge even with context, comprising $59.8\%$ of iterative+ctx and $55.8\%$ of Autocomp samples.

\subsection{Comparison Against Hand-Tuned Pallas}
\label{sec:results-handtuned}

Eight of the 17 priority operators ship with hand-optimized Pallas implementations from the upstream JAX Pallas ops library (Tokamax), with block sizes tuned via grid search on TPU v6e. We measured these kernels through the same JAXBench harness and report per-kernel speedups in Table~\ref{tab:handtuned}. Tokamax reaches a $2.08\times$ floor-1 geomean over the 8 kernels (the one sub-$1\times$ entry is a Splash Attention variant tuned for a different sparsity pattern than our dense-mask workload). Autocomp's $1.60\times$ geomean on the same subset reaches roughly $77\%$ of the hand-tuned geomean, exceeding Tokamax on 2 kernels and landing within $68$--$91\%$ on 4 more, but failing correctness on the paged- and ragged-attention operators where Tokamax's hand-written scheduling matters most. Iterative+context with the same sample budget reaches $1.34\times$.

\begin{table}[h]
\centering
\small
\caption{Speedups on the 8 priority kernels with hand-tuned Pallas references. Agents use a 144-sample Gemini 3 Flash budget. ``---'' denotes no correct kernel. Bold marks the best method per row.}
\label{tab:handtuned}
\vspace{0.5em}
\begin{tabular}{lrrrrrr}
\toprule
Benchmark & XLA (ms) & HT (ms) & Hand-tuned & Autocomp & Iter+ctx & Iter \\
\midrule
Flash Attention        & 25.21  & 6.45   & \textbf{3.91$\times$} & 2.67$\times$ & 0.46$\times$ & --- \\
GQA Attention          & 51.13  & 14.59  & \textbf{3.50$\times$} & 2.63$\times$ & 2.63$\times$ & --- \\
MLA Attention          & 20.32  & 21.72  & 0.94$\times$ & 0.85$\times$ & \textbf{1.54$\times$} & --- \\
Sparse Attention       & 5.95   & 6.89   & 0.86$\times$ & \textbf{2.81$\times$} & 2.54$\times$ & --- \\
Paged Attention        & 7.82   & 3.41   & \textbf{2.29$\times$} & --- & 0.61$\times$ & --- \\
Ragged Paged Attention & 18.70  & 2.71   & \textbf{6.91$\times$} & --- & --- & --- \\
GEMM                   & 5.36   & 5.59   & 0.96$\times$ & 0.75$\times$ & \textbf{0.97$\times$} & 0.59$\times$ \\
Megablox GMM      
     & 3.26   & 2.01   & 1.62$\times$ & \textbf{2.21$\times$} & 0.27$\times$ & 0.52$\times$ \\
\midrule
Geomean (floor 1$\times$) &   &   & \textbf{2.08$\times$} & 1.60$\times$ & 1.34$\times$ & 1.00$\times$ \\
\bottomrule
\end{tabular}
\end{table}

\subsection{Model Capacity Ablation (Gemini 3.1 Pro)}
\label{sec:results-ablation}

\begin{figure}[h]
\centering
 \includegraphics[width=\linewidth]{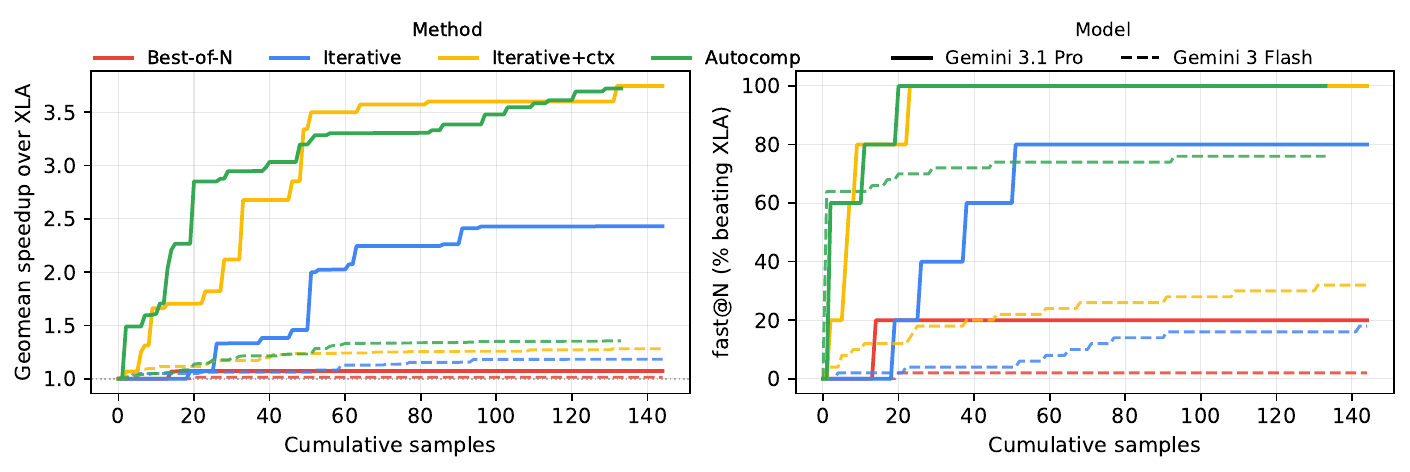}
\caption{Model-capacity ablation on the 5-kernel subset (RMSNorm, Flex Attention, RetNet Retention, Mamba-2 SSD, Flash Attention): speedup over XLA (left) and fast$_1$@$N$ (right) as a function of cumulative samples. Solid lines are Gemini 3.1 Pro; dashed lines are Gemini 3 Flash.}
\label{fig:trajectory}
\end{figure}

To measure the effect of model capacity, we rerun all four methods on a 5-kernel subset using Gemini 3.1 Pro, limited in scope due to Pro's substantially higher per-sample cost. The Flash curves in this ablation are remeasured on the same subset so that Flash and Pro are compared on identical benchmarks and sample budgets. Pro lifts every method, with the largest gains on iterative refinement, iterative+context, and Autocomp. Iterative jumps from $1.07\times$ to $2.43\times$, iterative+context from $1.59\times$ to $3.82\times$, and Autocomp from $2.35\times$ to $3.79\times$ (Figure~\ref{fig:trajectory}). Pro iterative+context and Autocomp land correct Pallas on all 5 benchmarks, whereas Flash Autocomp still fails on Mamba-2 SSD.

Plain iterative refinement benefits disproportionately from model scaling ($1.07\times \to 2.43\times$) purely from better Pallas priors, and once both context and capacity are engaged, iterative+context and Autocomp converge to essentially the same geomean ($3.82\times$ vs.\ $3.79\times$). This is unlike the full Flash suite where Autocomp leads by a wide margin, since once Pro produces correct seeds without extensive debugging, iterative+context's samples go toward optimization rather than correctness recovery. On the two kernels where Autocomp Pro trailed iterative+context (Mamba-2 SSD and Flash Attention), Autocomp's best latency was still improving monotonically through the final iteration, so the per-kernel ordering reflects exploration depth at a fixed budget rather than beam-search convergence. We treat this narrowing as a preliminary observation given the 5-kernel scope.

\section{Discussion and Future Work}

Compiler, runtime, and profiler feedback alone are insufficient to drive performance. Iterative refinement without TPU-specific documentation achieves only $16.9\%$ per-sample correctness on the 5-kernel subset with Gemini 3.1 Pro, and $1.2\%$ with Flash. The Pallas constraints that govern correctness, including lexicographic grid traversal, VMEM/SMEM/HBM placement, $(8, 128)$ block divisibility, and prefetch scheduling, are absent from error messages, leaving feedback loops nothing to iterate against.

Using a larger model helps, but not as much as adding useful context. Autocomp-generated documentation injected into the iterative loop raises per-sample correctness from $5.8\%$ to $37.3\%$ across the full Flash suite (Section~\ref{sec:results-full}), a much larger jump than the Flash-to-Pro gain at fixed context. For Pallas and other languages underrepresented in training data, the correctness bottleneck is information rather than reasoning, and written documentation supplies most of what is missing.

Once correctness is achieved, performance depends on how the sample budget is spent. Iterative refinement with only documentation solves \emph{more} benchmarks than Autocomp on the full Flash suite (48/50 vs.\ 45/50) but rarely pushes past the first correct kernel, finishing at $1.28\times$ geomean. Autocomp's translate/optimize split shifts the beam into optimization as soon as a correct seed exists, reaching $1.36\times$. Search structure, not just documentation, converts correctness into performance, though the Pro ablation shows this gap narrows once the model is strong enough to produce correct seeds without extensive debugging (Section~\ref{sec:results-ablation}).

\paragraph{Limitations.}
All 50 workloads run on a single TPU v6e chip, and multi-chip sharding and collective communication are out of scope. Hand-tuned Pallas references are available for 8 of the 17 priority operators and serve as an aspirational upper bound. The Gemini 3.1 Pro evaluation is restricted to a 5-kernel subset and is reported as a controlled model-capacity ablation rather than a primary result.

\paragraph{Future work.}
The most consequential extension is supporting multi-TPU kernel evaluation and optimization. Production training and inference workloads almost universally run across many chips, where end-to-end performance depends on sharding strategy, collective scheduling, and the overlap of communication with compute. Future work will extend \textsc{JAXBench} along the parallelism axis, covering tensor-parallel, pipeline-parallel, and expert-parallel sharding regimes on multi-chip TPU pods. A benchmark with these characteristics would evaluate whether the documentation-versus-search separation we observe generalizes to collective primitives (\texttt{shard\_map}, \texttt{psum\_scatter}, \texttt{all\_to\_all}, asynchronous collectives, and overlapping pipelines), which are even more sparsely represented in pretraining data than Pallas itself. More broadly, scaling AI-for-systems methods from single-device benchmarks to real-world deployment surfaces recurring challenges around topology-aware scheduling, communication overhead, and reproducibility of measured speedups~\cite{tschand2026genaisystemsrecurringchallenges}.

Beyond multi-chip, the information-gap result motivates fine-tuning or reinforcement learning from execution feedback on curated Pallas corpora, for which \textsc{JAXBench} correctness and latency signals are usable as rewards. The tier structure in our breakdown also suggests hybrid controllers that route between iterative refinement and beam-search agents on a per-benchmark basis. We release the benchmark, baselines, and harness to support these directions.

\section{Conclusion}
Hand-optimized accelerator kernels remain a persistent bottleneck for ML systems, and the pool of engineers fluent in both ML and low-level systems programming cannot keep pace with new architectures and hardware revisions. Recent benchmarks show that LLMs can begin to fill this gap on CUDA and Triton, but the same machinery fails on TPUs because Pallas is underrepresented in training data. \textsc{JAXBench} helps close the measurement gap with 50 production-scale JAX workloads, 8 hand-tuned Pallas references with grid-searched block sizes, and a profiler-grounded evaluation harness on TPU v6e.

Across the suite, automated methods make real progress but still trail expert kernels. Methods conditioned on documentation, like iterative refinement and Autocomp's beam search, beat the baseline XLA performance on many workloadse. The dominant lever on niche backends is target-specific context rather than model scale or feedback alone, and search structure is what converts correctness into performance once that information gap is closed. We release the suite, baselines, and harness to support evaluation for current methods and set a foundation for the multi-chip and training-time work to come.

\section*{Acknowledgements}
We thank Alain Hamel, Newfel Harrat, Steven Ingram, Mehrdad Khani, Gerson Kroiz, and Tomas Pfister for their helpful feedback and support throughout this work.

\bibliographystyle{plainnat}
\bibliography{jaxbench}

\appendix

\section{Per-Benchmark Results}
\label{app:per-benchmark}

\begin{figure}[h]
\centering
\includegraphics[width=0.6\linewidth]{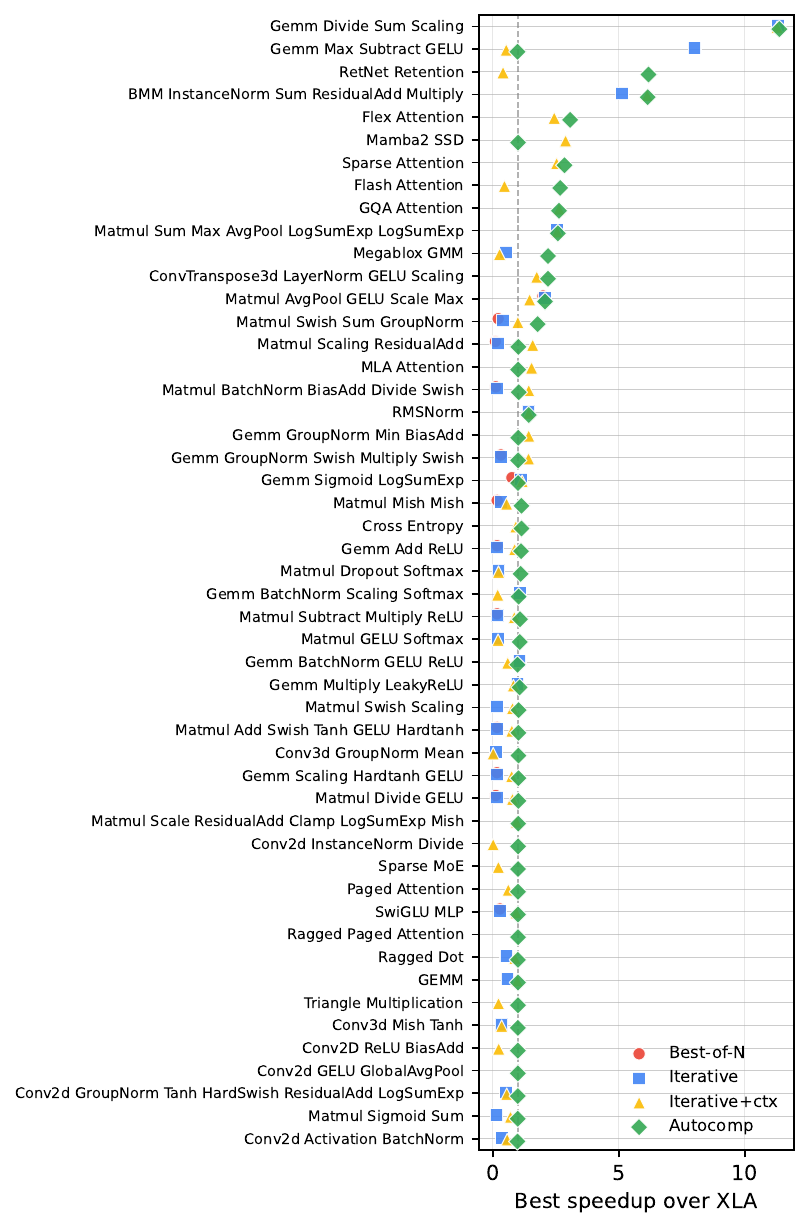}
\caption{Best speedup over the XLA baseline for each benchmark and method. Benchmarks without a marker had no correct implementation within the sample budget.}
\label{fig:per-benchmark}
\end{figure}

Figure~\ref{fig:per-benchmark} shows the best speedup achieved by each method on every benchmark in the 50-workload JAXBench suite with Gemini 3 Flash.
Benchmarks are sorted by the best speedup achieved by any method, and missing markers indicate the method produced no correct implementation within its sample budget.

Table~\ref{tab:pro-per-benchmark} reports the corresponding per-benchmark best speedups for the Gemini 3.1 Pro ablation on the 5-kernel subset (Section~\ref{sec:results-ablation}).

\begin{table}[h]
\centering
\small
\caption{Per-benchmark best speedup over XLA on the 5-kernel subset with Gemini 3.1 Pro. ``---'' denotes no correct kernel within the 144-sample budget. Bold marks the best method per row. Geomean floors per-kernel speedups at $1\times$ before aggregating.}
\label{tab:pro-per-benchmark}
\vspace{0.5em}
\begin{tabular}{lrrrrr}
\toprule
Benchmark & XLA (ms) & Best-of-$N$ & Iter & Iter+ctx & Autocomp \\
\midrule
RMSNorm            &  1.24 & 1.42$\times$ & 1.42$\times$ & 1.43$\times$ & \textbf{1.44$\times$} \\
Flex Attention     & 36.81 & ---          & 3.99$\times$ & 3.46$\times$ & \textbf{4.01$\times$} \\
RetNet Retention   & 13.04 & ---          & 4.85$\times$ & 6.40$\times$ & \textbf{9.62$\times$} \\
Mamba-2 SSD        & 29.59 & ---          & ---          & \textbf{5.66$\times$} & 4.04$\times$ \\
Flash Attention    & 25.38 & ---          & 3.07$\times$ & \textbf{4.54$\times$} & 3.51$\times$ \\
\midrule
Geomean (floor 1$\times$) & & 1.07$\times$ & 2.43$\times$ & \textbf{3.82$\times$} & 3.79$\times$ \\
\bottomrule
\end{tabular}
\end{table}

\section{Baseline Prompts and Agent Context}
\label{app:prompts}

This appendix documents the exact prompt construction for each evaluated method so the setup can be reproduced from the paper text alone.

\subsection{Best-of-$N$ and Iterative Refinement}

Both baselines use the same short system preamble:

\begin{quote}\small\ttfamily
You are optimizing a JAX kernel for TPU v6e (Trillium) using the Pallas programming model (\texttt{jax.experimental.pallas}).\\
Target hardware: TPU v6e. VMEM = 128 MiB, 8 TensorCores per chip, peak $\approx$918 TFLOPS bf16, HBM $\approx$1640 GB/s.\\
Rules for every turn: (1) keep the public \texttt{workload(*inputs)} signature unchanged; (2) output a single complete Python file inside one \mbox{```python\,...\,```} block; (3) do not include explanatory prose outside the code block.\\
Strategy: if you do not yet have a correct Pallas implementation, your first priority is to produce a correct, straightforward translation even if it isn't faster than the XLA baseline. Only after you have a correct Pallas kernel should you focus on optimizing it for speed.
\end{quote}

\paragraph{Best-of-$N$.} Each of the $N$ samples is an independent completion of a prompt consisting of the preamble above, a single line \texttt{Benchmark: <prob\_id>}, a one-sentence task description (``Below is the XLA/JAX reference implementation. Rewrite it as a Pallas kernel that produces identical outputs and runs faster.''), and the JAX source wrapped in a Python code block. No feedback is provided.

\paragraph{Iterative refinement.} The first turn of each chain uses the same prompt as Best-of-$N$. Each subsequent turn within a chain appends the following feedback block, then asks the model for a new complete Pallas implementation:

\begin{itemize}
    \item If the previous attempt was correct: ``Previous attempt was correct. Latency: $<$x$>$ ms (XLA baseline: $<$y$>$ ms, speedup: $<$s$>\times$). Best-so-far in this chain: $<$z$>$ ms.''
    \item If the previous attempt was incorrect: ``Previous attempt was incorrect or failed to run.'' followed by the last 40 lines of the runner's stdout.
    \item A profiler summary (top operators by time) from the previous run when available.
    \item The previous implementation's full source, wrapped in a Python code block.
\end{itemize}

\noindent Chains run independently and never share context, and 18 chains of 8 turns give a total of 144 samples per benchmark.

\paragraph{Iterative refinement with Autocomp context.} The \emph{iterative+context} variant uses the same chain/turn/feedback layout as plain iterative refinement (18 chains $\times$ 8 turns, same feedback block on correctness or failure, same final-code echo) with one change. Before every prompt we prepend the static portion of Autocomp's agent context, namely the hardware architecture summary, the per-benchmark-selected Pallas API reference, the selected code examples, and the rules block (see next subsection). The per-benchmark category and example selections are made by the same LLM yes/no filter Autocomp uses and are cached per benchmark, so a chain's eight turns see the same context menu. The variant does not use Autocomp's beam search or translate/optimize phase split, and is purely a context injection into the iterative baseline.

\subsection{Autocomp Context Artifacts}

We passed four JAX and Cloud TPU documentation sources into Autocomp's Agent Builder pipeline:
\begin{itemize}
    \item JAX Pallas guide: \url{https://docs.jax.dev/en/latest/pallas/index.html}
    \item JAX Pallas TPU guide: \url{https://docs.jax.dev/en/latest/pallas/tpu/index.html}
    \item \texttt{jax.experimental.pallas.tpu} API reference: \url{https://docs.jax.dev/en/latest/jax.experimental.pallas.tpu.html}
    \item Cloud TPU documentation: \url{https://docs.cloud.google.com/tpu/docs/}
\end{itemize}
Each source is crawled with \texttt{max\_depth=2} and \texttt{max\_pages=250}. An LLM-driven routing-and-synthesis pass then distills the crawled content into four textual artifacts that together form the static portion of every agent prompt:

\begin{itemize}
    \item \textbf{Hardware architecture summary}~(\,$\approx$8~KB): a single-pass LLM summary of TPU v6e characteristics (memory hierarchy, compute units, bandwidth and capacity numbers, and programming-model constraints).
    \item \textbf{Pallas API / ISA reference}~(\,$\approx$115~KB): per-API entries (signature, description, parameter semantics, inline usage snippets) extracted verbatim from the source documentation and grouped into functional categories. At prompt time, an LLM yes/no filter selects which categories are relevant to the current kernel and caches the selection per-benchmark.
    \item \textbf{Annotated code examples}~(\,$\approx$30~KB): representative Pallas kernels drawn from the ingested documentation, each accompanied by a one-line summary. Examples are selected per-benchmark via a parallel LLM yes/no filter.
    \item \textbf{Rules}~(\,$\approx$6~KB): correctness constraints (e.g., block-shape divisibility, memory-space annotations, aliasing rules) extracted as imperative bullet points and appended to every planning and coding prompt.
\end{itemize}

\noindent During beam search, each candidate-generation prompt concatenates the architecture summary, the per-benchmark-selected API reference, any selected code examples, the candidate's current code (plus its latest score and any hardware-counter feedback), and the rules block, before asking the model for either a plan or a complete optimized kernel. The translate and optimize phases use the same scaffold but draw from separate menus of strategies.

\end{document}